\documentclass[letterpaper]{article} 
\usepackage[]{aaai24}  
\usepackage{times}  
\usepackage{helvet}  
\usepackage{courier}  
\usepackage[hyphens]{url}  
\usepackage{graphicx} 
\usepackage{natbib}  
\usepackage{caption} 
\usepackage{algorithm}
\usepackage{algorithmic}

\newcommand{\ctext}[1]{\raise0.2ex\hbox{\textcircled{\scriptsize{#1}}}}
\usepackage{multirow}
\usepackage{color, colortbl}
\definecolor{Gray}{gray}{0.8}
\usepackage{amsmath}
\usepackage{footmisc}
\usepackage{longtable}
\usepackage{multicol}
\usepackage{newfloat}
\usepackage{listings}
\usepackage{here}
\DeclareCaptionStyle{ruled}{labelfont=normalfont,labelsep=colon,strut=off} 
\floatstyle{ruled}
\newfloat{listing}{tb}{lst}{}
\floatname{listing}{Listing}
\title{OUTFOX: LLM-Generated Essay Detection Through In-Context Learning \\with Adversarially Generated Examples}
\author {
    Ryuto Koike\textsuperscript{\rm 1},
    Masahiro Kaneko\textsuperscript{\rm 2,1},
    Naoaki Okazaki\textsuperscript{\rm 1}
}
\affiliations {
    \textsuperscript{\rm 1}Tokyo Institute of Technology\\
    \textsuperscript{\rm 2}MBZUAI\\
    ryuto.koike@nlp.c.titech.ac.jp, masahiro.kaneko@mbzuai.ac.ae, okazaki@c.titech.ac.jp
}

\begin{document}

\maketitle

\begin{abstract}
Large Language Models (LLMs) have achieved human-level fluency in text generation, making it difficult to distinguish between human-written and LLM-generated texts. 
This poses a growing risk of misuse of LLMs and demands the development of detectors to identify LLM-generated texts.
However, existing detectors lack robustness against attacks: they degrade detection accuracy by simply paraphrasing LLM-generated texts.
Furthermore, a malicious user might attempt to deliberately evade the detectors based on detection results, but this has not been assumed in previous studies.
In this paper, we propose \textbf{OUTFOX}, a framework that improves the robustness of LLM-generated-text detectors by allowing both the detector and the attacker to consider each other's output.
In this framework, the attacker uses the detector's prediction labels as examples for in-context learning and adversarially generates essays that are harder to detect, while the detector uses the adversarially generated essays as examples for in-context learning to learn to detect essays from a strong attacker.
Experiments in the domain of student essays show that the proposed detector improves the detection performance on the attacker-generated texts by up to +41.3 points F1-score. 
Furthermore, the proposed detector shows a state-of-the-art detection performance: up to 96.9 points F1-score, beating existing detectors on non-attacked texts.
Finally, the proposed attacker drastically degrades the performance of detectors by up to -57.0 points F1-score, massively outperforming the baseline paraphrasing method for evading detection.
\end{abstract}

\begin{figure}[t]
 \begin{center}
  \centering\includegraphics[width=\columnwidth]{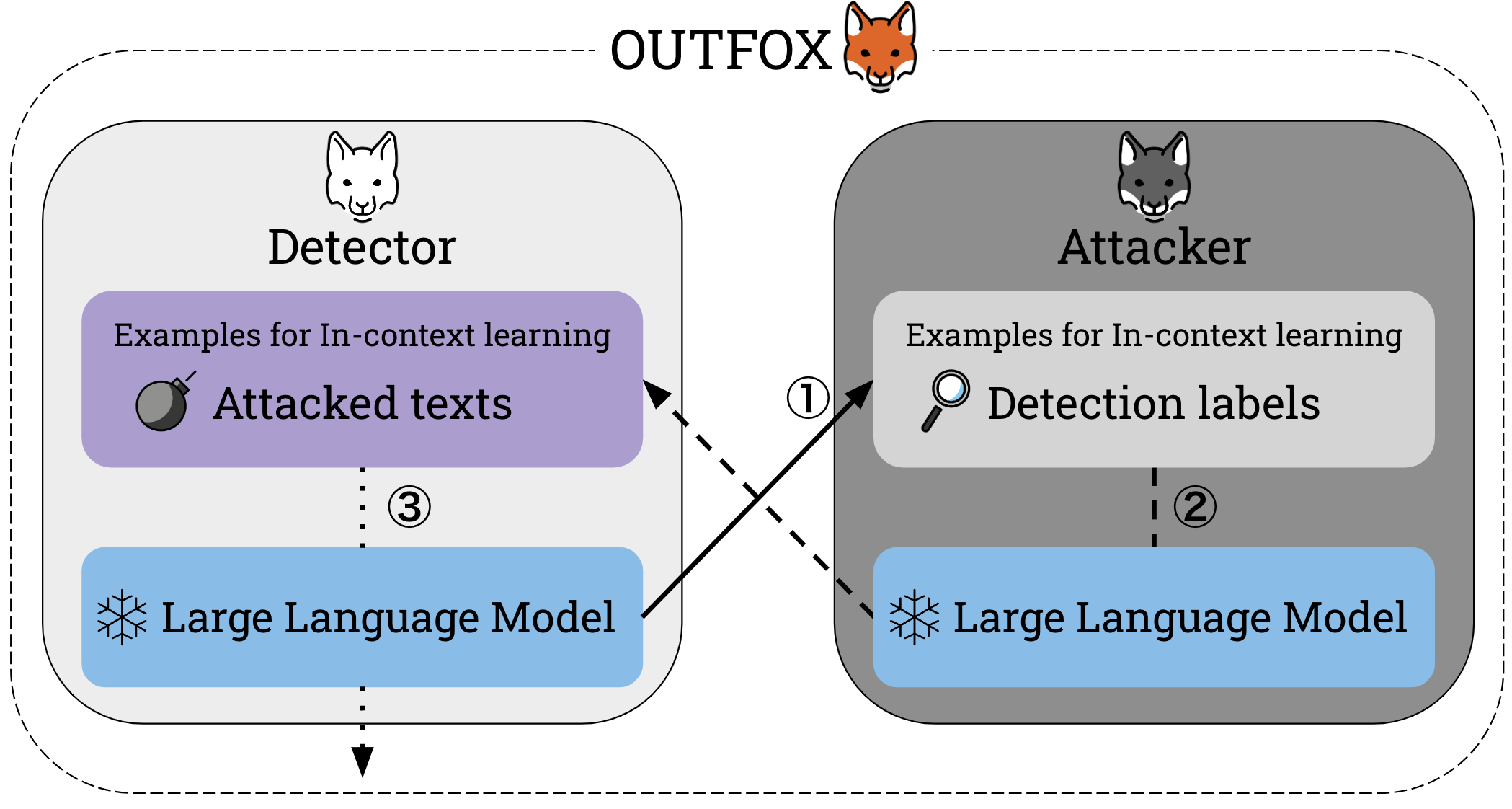}
  \caption{
  In our OUTFOX framework, there are three steps.
  Step \ctext{1}: The detector outputs prediction labels to texts in a training set. 
  Step \ctext{2}: The attacker uses the detector's prediction labels as examples for in-context learning to generate more sophisticated attacks against a training set.
  Step \ctext{3}: The detector uses these adversarially generated texts by a strong attacker to detect texts in a test set.}
  \vspace*{-0.6cm}
  \label{fig:outfox}
 \end{center}
\end{figure}

\section{Introduction}
LLMs, characterized by their enormous model size and vast training data, have demonstrated emergent abilities with impressive performance across various tasks \citep{wei2022emergent}. 
These abilities include a high degree of language comprehension, fluent generation, and the capacity to handle tasks unseen during training through in-context learning \citep{brown2020language,ouyang2022training,sanh2022multitask}.

Despite these successes, there are growing concerns about the potential misuse of LLMs.
A notable example is in education, where students might copy and paste text generated by LLMs, such as ChatGPT \citep{chatgpt}, for their assignments. 
This concern has led to the development of detectors designed to identify LLM-generated text with promising detection performance \citep{kirchenbauer2023watermark,mitchell2023detectgpt,aitextclassifier,tang2023science}.

Unfortunately, existing detectors often perform poorly against simple attacks (e.g., paraphrasing), as highlighted by recent studies \citep{sadasivan2023aigenerated,krishna2023paraphrasing}.
A recent survey called for developing robust detection methods against other potential attacks designed to deceive the detectors \citep{tang2023science}.
Given the human-like generative abilities of LLMs, there's the unexplored risk that malicious users might exploit LLMs to create texts specifically designed to evade detection.

Motivated by this need, we propose \textbf{OUTFOX}, a novel framework designed to enhance the robustness and applicability of LLM-generated text detectors. 
As illustrated in Figure \ref{fig:outfox}, OUTFOX introduces an approach where both the detector and the attacker learn from each other's output. The attacker uses the detector's predictions as examples for in-context learning to generate more sophisticated attacks. In contrast, the detector uses these adversarially generated texts as examples for in-context learning to improve its detection abilities against a strong attacker.
To validate our approach in a realistic domain, we create a dataset of native-speaker student essay writing to detect LLM-generated essays. 
Our dataset contains 15,400 triplets of essay problem statements, student-written essays, and LLM-generated essays.

Experiments show that our OUTFOX detector substantially improves the detection performance on the attacker-generated texts by up to +41.3 points F1-score compared to without considering attacks.
This result empirically suggests that LLMs, especially ChatGPT, might learn implicit differences between non-attacked and attacked texts via in-context examples.
Interestingly, our OUTFOX detector performs consistently as well or even better on non-attacked texts, resulting in the state-of-the-art detection performance of up to 96.9 points F1-score on non-attacked texts.
This result demonstrates that considering attacks in our detector has little negative effect on detection performance on non-attacked texts.
Furthermore, our OUTFOX attacker can drastically degrade the performance of detectors, with a decrease of up to -57.0 points F1-score, which massively outperforms the baseline paraphrasing method for evading detection.
In our analysis, we explore the semantic differences between non-attacked and attacked essays. The analysis reveals that our attacker-generated essays are semantically closer to human-written essays than non-attacked essays, leading to success in such effective attacking. For reproducibility, we release our code and dataset.\footnote{\url{https://github.com/ryuryukke/OUTFOX}}

\begin{figure*}[t]
 \begin{center}
  \includegraphics[width=\textwidth]{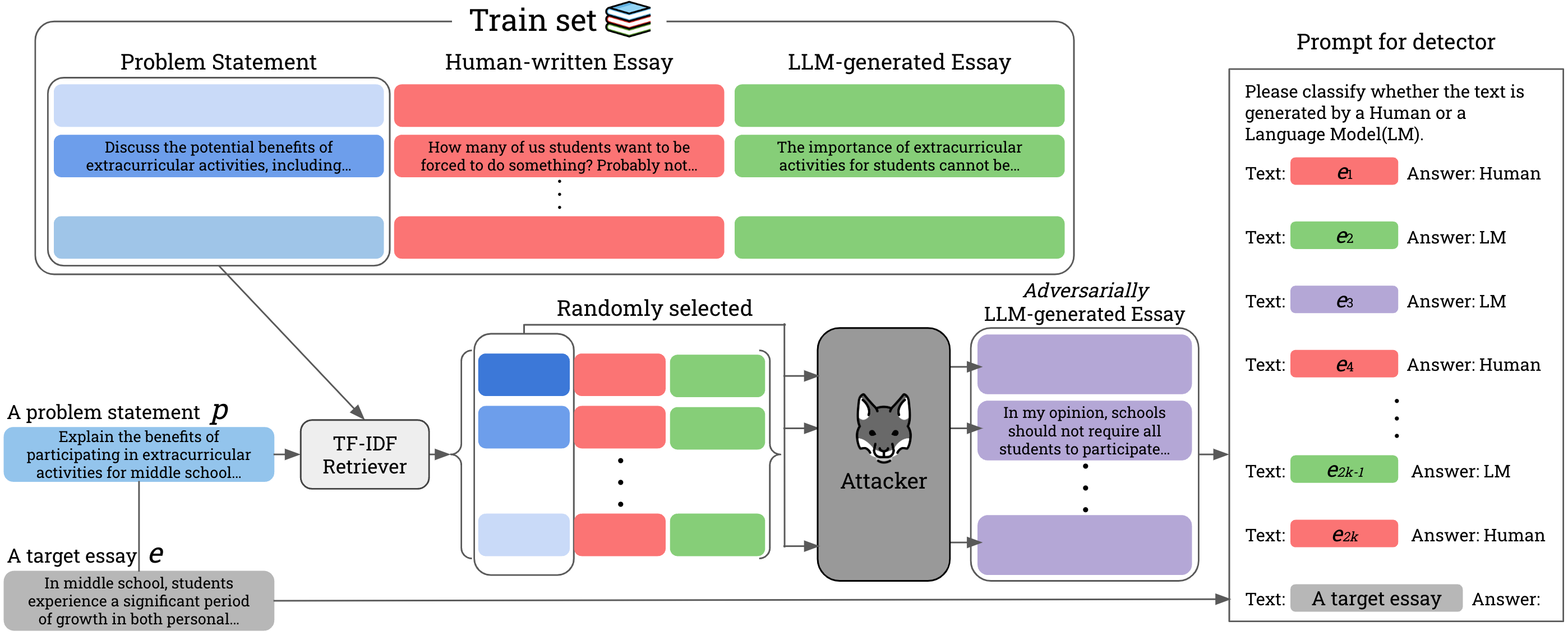}
  \caption{An illustration of our OUTFOX detector: The detector utilizes the adversarially generated essays as examples for in-context learning to learn to detect essays from our OUTFOX attacker.}
  \label{fig:detector}
 \end{center}
\end{figure*}

\begin{figure*}[t]
 \begin{center}
  \includegraphics[width=\textwidth]{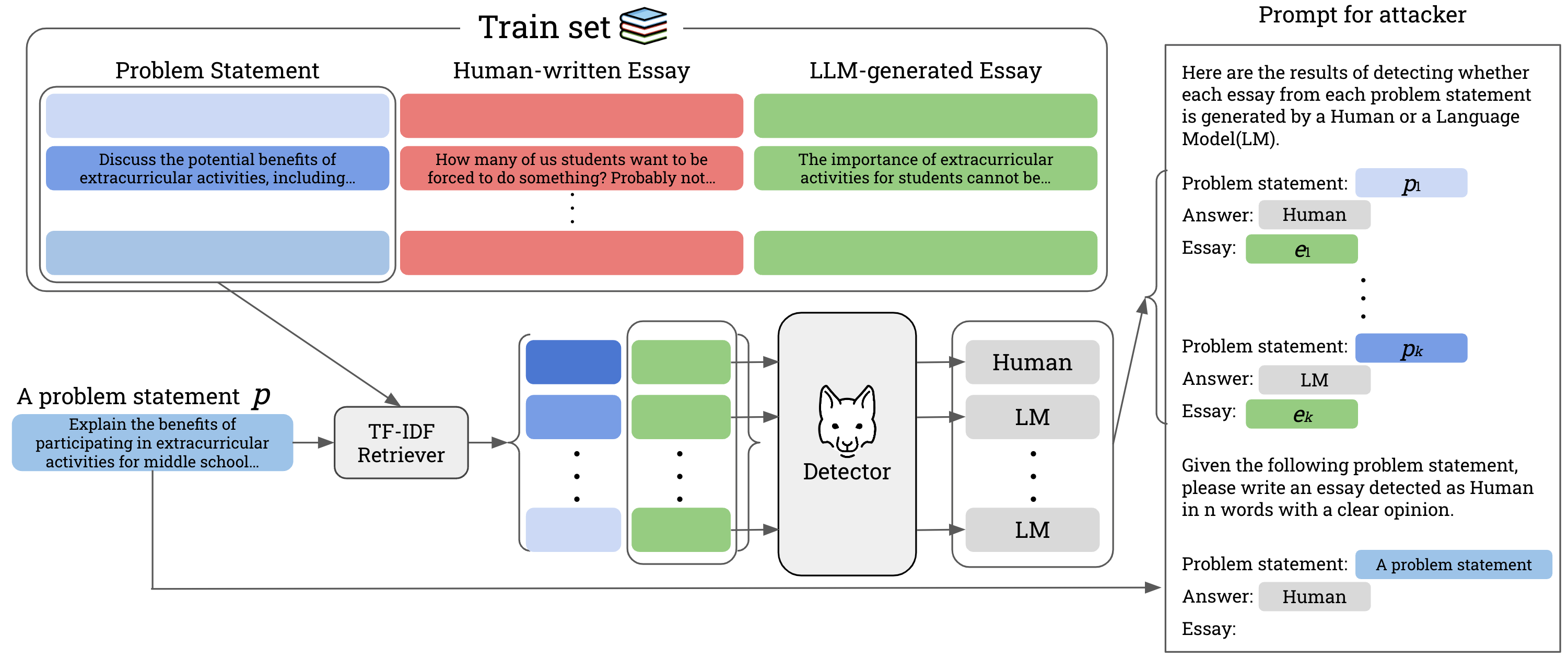}
  \caption{An illustration of our OUTFOX attacker: The attacker considers our OUTFOX detector's prediction labels as examples for in-context learning and adversarially generates essays that are harder to detect.}
  \label{fig:attacker}
 \end{center}
\end{figure*}

\section{Related Work}

\paragraph{LLM-Generated Text Detection}
\label{baselines}
Tackling the malicious uses of LLMs, recent studies have proposed detectors to identify LLM-generated texts.
These detectors can be mainly categorized into watermarking algorithms, statistical outlier detection methods, and supervised classifiers.
Watermarking algorithms use token-level secret markers in texts that humans cannot recognize for detection \citep{kirchenbauer2023watermark}.
In order to embed the markers into texts, the probabilities of selected tokens by a hash function are modified to be higher in text generation at each time step.
Our work focuses not on the watermark-enhanced LMs, but on LLMs that are openly used in our daily life---hence our chosen domain of student essays.
Statistical outlier detection methods exploit statistical differences in linguistic features between human-written and LLM-generated texts. These includes n-gram frequencies \citep{badaskar-etal-2008-identifying}, entropy \citep{ent08}, perplexity \citep{bere16}, token log probabilities \citep{solaiman2019release}, token ranks \citep{ippolito-etal-2020-automatic}, and negative curvature regions of the model’s log probability \citep{mitchell2023detectgpt}. 
Supervised classifiers are models specifically trained to discriminate human-written and LLM-generated texts with labels. The classifiers range from classical methods \citep{ippolito-etal-2020-automatic,crothers2023machine} to neural-based methods \citep{solaiman2019release,bakhtin2019real,uchendu-etal-2020-authorship,rodriguez-etal-2022-cross,guo2023close}. 

\paragraph{Detection for Assessing Academic Dishonesty}
In the educational context, LLM-based services (e.g., ChatGPT and Bard \cite{bard}) have the potential to help students with their writing. However, many schools consider these services as academic dishonesty because students can misuse these tools to cheat on assignments \citep{educatorconsiderations4chatgpt}.
Based on this, recent studies have investigated LLM-generated text detection for student assignments, such as argumentative essays \citep{liu2023argugpt} and university-level course problems \citep{ibrahim2023perception,vasilatos2023howkgpt}.
Specifically, \citet{liu2023argugpt} targeted the essay domain written by non-native English learners. However, considering the human-level generative capabilities of LLMs, it can be more difficult to classify LLM-generated and native-written essays than non-native-written ones. Thus, as a more challenging setting, we create a dataset to distinguish native-student-written from LLM-generated essays.

\noindent
\paragraph{Attacking LLM-Generated Text Detection}
\label{attacking}
Most recent studies have reported the effectiveness of paraphrasing attacks, where existing detectors experience a large loss in accuracy when given a paraphrased version of an LLM-generated text \citep{sadasivan2023aigenerated,krishna2023paraphrasing}.
For instance, \citet{krishna2023paraphrasing} proposed DIPPER, the 11B document-level paraphraser, controlling output diversity on vocabulary and content re-ordering.
Considering the human-level generative abilities of LLMs, malicious users might instruct LLMs to adversarially generate texts based on detection results, but this has not been explored in prior work.

\noindent
\paragraph{Defense against Attacking LLM-Generated Text Detection}
The existence of these attacking methods, in turn, poses a need for a defense against such attacks. There are few studies specifically about defending against the attacks. \citet{krishna2023paraphrasing} proposed a retrieval-based defense that detects a text that is semantically similar to one of the LLM-generated responses in an API database.
However, it needs active actions by API providers, and based on the nature of the method, a false positive rate can be higher with a larger database.
In addition, \citet{sadasivan2023aigenerated} recently showed that the retrieval-based defense method is vulnerable against recursive paraphrasing. The detection accuracy significantly drops to 25\% after 5 rounds of paraphrasing by \citet{krishna2023paraphrasing}'s paraphraser.

In recent concurrent work by \citet{hu2023radar}, they proposed RADAR, a framework that jointly trains a robust LLM-generated-text classifier against the paraphrasing attack via adversarial learning.
In our OUTFOX framework, the detector and the attacker can learn from each other's adversarially generated output via in-context learning, without any parameter updates.
Thus, the detector in our framework can handle new attacks by simply adding these to in-context examples without additional fine-tuning.
Additionally, the attacker in our framework is not limited to the paraphrasing attack but is directly designed to deceive the detector.

\section{OUTFOX Framework}
\label{methodlogy}

\subsection{Task Formulation}
Our work focuses on distinguishing LLM-generated essays from human-written essays. We assume that a training instance for detecting LLM-generated essays consists of a triplet of an essay-problem statement, a human-written essay, and an LLM-generated essay.
In our framework, the attacker adversarially generates \textit{attacked} text, designed to fool the detector, while the detector tries to learn from examples to better distinguish LLM-generated and human-written text. Both the detector and attacker are able to \textit{consider} each other's outputs by including examples in their input prompts for in-context learning. The detector can consider essay-label pairs that can consist of human-written, \textit{attacked} (adversarially-generated by an attacker), and \textit{non-attacked} (generated by an LLM, but not adversarially) essays by including them in its input prompt. On the other hand, the attacker can consider examples of non-attacked text and the detector's label predictions in its input prompt.

\subsection{The OUTFOX Detector}
\label{detector}
Figure \ref{fig:detector} illustrates the outline of our OUTFOX detector.
Given a target essay $e$ paired with a problem statement $p$, we retrieve the top-$k$ problem statements that are semantically close\footnote{We follow the setting of \citet{prabhumoye2022fewshot}, which found that leveraging semantically close examples is effective for in-context learning.
We use the vector space of all problem statements computed by Term Frequency-Inverse Document Frequency (TF-IDF) implemented in scikit-learn:\linebreak\url{https://tinyurl.com/scikitlearn-TF-IDF-vectorizer}. The closeness is computed by $(1 - s)$, where $s$ presents the cosine similarity of the vectors of two problem statements.\label{similarity}} to the target problem statement $p$ from the training set, together with human-written and LLM-generated essays associated with the retrieved problem statements.
To generate attacks to be considered in detection, we randomly select $j$ out of the $k$ problem statements ($0 \leq j \leq k$).
Consequently, our OUTFOX attacker adversarially generates an essay for each selected problem statement.
In this way, we obtain $k$ human-written, $(k-j)$ LLM-generated, and $j$ attacked essays and construct a mixture of these essays with labels $R_{\rm det}=\{(e_i, l_i)\}_{i=1}^{2k}$.
Here, $l_i$ is \texttt{Human} when the essay $e_i$ is written by a human, or \texttt{LM} when the essay $e_i$ is written by either the LLM or attacker.
Tagging attacker-generated texts with \texttt{LM} label encourages our detector to learn the implicit characteristics of attacker-generated texts for detection.
Finally, the detector predicts a label $\hat{l}$ for the target essay $e$ by using the LLM with the task instruction $I$ and $R_{\rm det}$ as a prompt,
\begin{equation}
\hat{l} = \underset{l \in \{ \text{\texttt{Human}}, \text{\texttt{LM}}\}}{\operatorname{argmax}} P\left(l \mid I, R_{\rm det}, e\right) ,
\end{equation}
\begin{equation}
R_{\rm det}=\{(e_i, l_i)\}_{i=1}^{2k} .
\end{equation}
Here, the task instruction $I$ is ``Please classify whether the text is generated by a human or a Language Model (LM).''
Each example in $R_{\rm det}$ is converted into the text format, ``Text: $e_i$ Answer: $l_i$.''
The target essay $e$ is converted into ``Text: $e$. Answer: '' so that the LLM predicts a label as a continuation of the prompt.


\subsection{The OUTFOX Attacker}
\label{attacker}
Figure \ref{fig:attacker} illustrates the workflow of our OUTFOX attacker. Unlike the paraphrasing attack, our attacker generates an essay from a problem statement so that it fools our detector. Given a problem statement $p$, we first retrieve the top-$k$ problem statements $p_{i}\;(i \in \{1, \dots, k\})$ that are semantically close\footref{similarity} to the problem statement $p$ from the training set, along with human-written and LLM-generated essays $e_i$ associated with the retrieved problem statements.
Then, our detector predicts a label $\hat{l}_i \in \{\texttt{Human}, \texttt{LM}\}$ on each retrieved LLM-generated essay $e_i$ using in-context examples from the training set.
At this time, our detector does not consider an attack.\footnote{Although our framework theoretically allows the detector and attacker to iteratively strengthen each other many times, we focus on only once.}
Consequently, we create a mixture of the retrieved problem statements, the predicted labels, and the retrieved LLM-generated essays $R_{\rm atk}=\{(p_i, \hat{l}_i, e_i)\}_{i=1}^{k}$.
Finally, our attacker adversarially generates an essay $a$ as a continuation for the prompt rendered from $I_{\rm d}$, $R_{\rm atk}$, the task instruction of essay-generation $I$, and the given problem statement $p$. We use an LLM to predict the next token $\hat{a}_t$ for the previously predicted tokens $\hat{a}_{1:t-1}$,
\begin{equation}
\hat{a}_t \approx \underset{a_t \in V}{\operatorname{argmax}} P\left(a_t \mid I_{\rm d}, R_{\rm atk}, I, p, \hat{a}_{1:t-1}\right) ,
\end{equation}
\begin{equation}
R_{\rm atk}=\{(p_i, \hat{l}_i, e_i)\}_{i=1}^{k} .
\end{equation}
Here, $V$ is the vocabulary of the LLM.
The description $I_{\rm d}$ is ``Here are the results of detecting whether each essay from each problem statement is generated by a Human or a Language Model (LM)''. Each example in $R_{\rm atk}$ is converted to the text format, ``Problem Statement: $p_{i}$. Answer: $\hat{l}_{i}$. Essay: $m_{i}$.''
The task instruction $I$ is ``Given the following problem statement, please write an essay detected as Human in $N$ words with a clear opinion''.
$N$ is the number of words in the human-written essay paired with the given problem statement $p$.
The problem statement $p$ is converted into ``Problem Statement: $p$. Answer: Human. Essay: '' so that the LLM adversarially generates an essay to fool the detector as a continuation of the prompt, including explicit \texttt{Human} label.

\begin{table*}[t]
\centering
\small
\setlength{\tabcolsep}{6pt} 
\renewcommand{\arraystretch}{1.05} 
\begin{tabular}{cccccc}\hline
\multirow{2}{*}{\textbf{Attacker}}& \multirow{2}{*}{\textbf{Detector}} & \multicolumn{4}{c}{\textbf{Metrics (\%) ↑}}\\\cline{3-6}
& & \textbf{HumanRec} & \textbf{MachineRec} & \textbf{AvgRec} & \textbf{F1}\\\hline
\multirow{3}{*}{DIPPER}&\multicolumn{1}{l}{w/o Attacks} & 98.6 & 66.2 & 82.4 & 79.0\\
&\multicolumn{1}{l}{w/\hspace{2.5mm}DIPPER} & 98.2 & 79.6 & \textbf{88.9} & \textbf{87.8} \\
&\multicolumn{1}{l}{w/\hspace{2.5mm}OUTFOX} & 97.8 & 72.4 & 85.1 & 82.9\\
\hline
\multirow{3}{*}{OUTFOX}&\multicolumn{1}{l}{w/o Attacks} & 98.8 & 24.8 & 61.8 & 39.4\\
&\multicolumn{1}{l}{w/\hspace{2.5mm}DIPPER} & 98.6 & 20.8 & 59.7 & 34.0 \\
&\multicolumn{1}{l}{w/\hspace{2.5mm}OUTFOX} & 97.2 & 69.6 & \textbf{83.4} & \textbf{80.7}\\
\hline
\end{tabular}
\caption{Comparison of the detection performances of our OUTFOX detector on attacked essays, with and without considering attacks: our OUTFOX attack and the DIPPER attack. The DIPPER paraphrases ChatGPT-generated essays for attacking. In the rows of ``w/o Attacks'', we show the detection performances of our detector, without considering attacks, on attacked essays by each attacker.}
\label{robustness_of_our_detector}
\end{table*}

\section{Constructing a Dataset to Detect LLM-Generated Essays}
\label{dataset}
We build a dataset for student essay writing, specifically to detect LLM-generated essays. There are already some datasets for the purpose of automatically scoring student-written essays \citep{asap-aes,feedback-prize-effectiveness}, but few of them have abundant essay-problem statement pairs. To get LLM-generated essays, we need essay problem statements.
We focus on the essay dataset of \citet{feedback-prize-effectiveness}, consisting of argumentative essays written by native students from 6th to 12th grade in the U.S.
Firstly, we instruct ChatGPT to generate a pseudo-problem-statement to mimic a setting where a student would produce the supplied essay. Afterward, we instruct an instruction-tuned LLM to generate an essay based on each generated problem statement.
For each LLM, our dataset contains 15,400 triplets of essay problem statements, student-written essays, and LLM-generated essays. In our evaluation, we split the dataset into three parts: train/validation/test with 14400/500/500 examples, respectively.
Besides the non-attacked LLM-generated essays, to evaluate each attacker in our experiments, we also build 500 attacked essays associated with problem statements in our test set for each attacker.

\section{Experiments and Results}
In our experiments, we investigate the following aspects: 
\begin{itemize}
\item How robust is our detector, considering attacks, against attacked texts? 
\item Does our detector, considering attacks, consistently perform well even on non-attacked texts? 
\item Is our attacker stronger than the previous paraphrasing attack approach?
\end{itemize}

\subsection{Overall Setup}
\label{experimental_setup}
\paragraph{Essay Generation Models}
To generate non-attacked essays, we instruct the instruction-tuned LMs: ChatGPT (gpt-3.5-turbo-0301), GPT-3.5 (text-davinci-003), and FLAN-T5-XXL\footnote{\url{https://huggingface.co/google/flan-t5-xxl}}.
In each, we set the $\mathsf{temperature}$ parameter to 1.3.

\paragraph{Evaluation Metrics and Dataset}
Area Under Receiver Operating Characteristic curve (AUROC) can be applied only to the detectors which output real number prediction scores.
Since our proposed detector outputs a binary label, we employ the F1-score on LLM-generated texts as our first metric.
Our second metric for detection performance is AvgRec, following \citet{li2023deepfake}.
AvgRec is the average of HumanRec and MachineRec.
In our evaluation, HumanRec is the recall for detecting Human-written texts, and MachineRec is the recall for detecting LLM-generated texts. 
We compute a classification threshold for each baseline detector on our validation set where the Youden Index\footnote{The Youden Index is the difference between True Positive Rate (TPR) and False Positive Rate (FPR). The cut-off point in the ROC curve, where the Youden Index is maximum, is the best trade-off between TPR and FPR.} \cite{1950youden} is maximum in the ROC curve.
Finally, we evaluate detectors with these metrics and thresholds on our test set: a mixture of 500 human-written and 500 non-attacked essays.
To evaluate detectors on attacked essays, we swap only LLM-generated essays from non-attacked to attacked ones. Additionally, the threshold for each detector is fixed on both non-attacked and attacked essays.

\begin{table*}[t]
\centering
\small
\setlength{\tabcolsep}{6pt} 
\renewcommand{\arraystretch}{1.05} 
\begin{tabular}{cccccc}\hline
\multirow{2}{*}{\textbf{Essay Generator}}& \multirow{2}{*}{\textbf{Detector}} & \multicolumn{4}{c}{\textbf{Metrics (\%) ↑}}\\\cline{3-6}
& & \textbf{HumanRec} & \textbf{MachineRec} & \textbf{AvgRec} & \textbf{F1}\\\hline
\multirow{3}{*}{ChatGPT}&\multicolumn{1}{l}{w/o Attacks} & 99.0 & 94.0 & \textbf{96.5} & \textbf{96.4}\\
&\multicolumn{1}{l}{w/\hspace{2.5mm}DIPPER} & 99.2 & 87.8 & 93.5 & 93.1 \\
&\multicolumn{1}{l}{w/\hspace{2.5mm}OUTFOX} & 97.8 & 92.4 & 95.1 & 95.0\\
\hline
\multirow{3}{*}{GPT-3.5}&\multicolumn{1}{l}{w/o Attacks} & 98.6 & 95.2 & \textbf{96.9} & 96.8\\
&\multicolumn{1}{l}{w/\hspace{2.5mm}DIPPER} & 98.8 & 92.4 & 95.6 & 95.5 \\
&\multicolumn{1}{l}{w/\hspace{2.5mm}OUTFOX} & 97.6 & 96.2 & \textbf{96.9} & \textbf{96.9}\\
\hline
\multirow{3}{*}{FLAN-T5-XXL}&\multicolumn{1}{l}{w/o Attacks} & 98.8 & 68.2 & 83.5 & 80.5\\
&\multicolumn{1}{l}{w/\hspace{2.5mm}DIPPER} & 99.2 & 72.0 & \textbf{85.6} & \textbf{83.3} \\
&\multicolumn{1}{l}{w/\hspace{2.5mm}OUTFOX} & 97.0 & 73.4 & 85.2 & 83.2\\
\hline
\end{tabular}
\caption{
Comparison of the detection performances of our OUTFOX detector on non-attacked essays, with and without considering the attacks: our OUTFOX attack and the DIPPER attack.
In the rows of ``w/o Attacks'', we show the detection performances of our detector, without considering attacks, on non-attacked essays.
}
\label{side_effect}
\end{table*}

\begin{table*}[t]
\centering
\small
\setlength{\tabcolsep}{6pt} 
\renewcommand{\arraystretch}{1.05} 
\begin{tabular}{cccccc}\hline
\multirow{2}{*}{\textbf{Detector}}& \multirow{2}{*}{\textbf{Attacker}} & \multicolumn{4}{c}{\textbf{Metrics (\%) ↓}}\\\cline{3-6}
& & \textbf{HumanRec} & \textbf{MachineRec} & \textbf{AvgRec} & \textbf{F1}\\\hline
\multirow{3}{*}{RoBERTa-base}&\multicolumn{1}{l}{Non-attacked} & 93.8 & 92.2 & 93.0 & 92.9\\
&\multicolumn{1}{l}{DIPPER} & 93.8 & 89.2 & 91.5 & 91.3 \\
&\multicolumn{1}{l}{OUTFOX} & 93.8 & 69.2 & \textbf{81.5} & \textbf{78.9}\\
\hline
\multirow{3}{*}{RoBERTa-large}&\multicolumn{1}{l}{Non-attacked} & 91.6 & 90.0 & 90.8 & 90.7\\
&\multicolumn{1}{l}{DIPPER} & 91.6 & 97.0 & 94.3 & 94.4 \\
&\multicolumn{1}{l}{OUTFOX} & 91.6 & 56.2 & \textbf{73.9} & \textbf{68.3}\\
\hline
\multirow{3}{*}{HC3 detector}&\multicolumn{1}{l}{Non-attacked} & 79.2 & 70.6 & 74.9 & 73.8\\
&\multicolumn{1}{l}{DIPPER} & 79.2 & 3.4 & 41.3 & 5.5 \\
&\multicolumn{1}{l}{OUTFOX} & 79.2 & 0.4 & \textbf{39.8} & \textbf{0.7}\\
\hline
\multirow{3}{*}{OUTFOX}&\multicolumn{1}{l}{Non-attacked} & 99.0 & 94.0 & 96.5 & 96.4\\
&\multicolumn{1}{l}{DIPPER} & 98.6 & 66.2 & 82.4 & 79.0 \\
&\multicolumn{1}{l}{OUTFOX} & 98.8 & 24.8 & \textbf{61.8} & \textbf{39.4}\\
\hline
\end{tabular}
\caption{
Comparison of the detection performances of the detectors on ChatGPT-generated essays, before and after being attacked: our OUTFOX attack and the DIPPER attack. In the rows of ``Non-attacked'', we show the detection performances of each detector on non-attacked essays.
}
\label{our_stronger_attacker}
\end{table*}

\paragraph{Detection Methods}
Our OUTFOX detector is based on ChatGPT (gpt-3.5-turbo-0301). We set the $\mathsf{temperature}$ and $\mathsf{top\_p}$ parameters to 0 in order to eliminate the randomness of our detection.
Our detector takes retrieved in-context examples for an essay.
In detection without considering attacks, the in-context examples are 5 human-written and 5 LLM-generated essays.
In detection with considering attacks, the in-context examples are 5 human-written, 3 attacked, and 2 LLM-generated essays.
Here, regardless of the essay generation models to be detected, our detector takes ChatGPT-generated essays as part of in-context examples.
As a comparison with prior work, we compare our detector to the following detectors, divided into two groups: statistical outlier approaches and supervised classifiers.
The first group covers Rank, LogRank, Log Probability, and DetectGPT (as explained in \S\ref{baselines}).
The second group contains OpenAI's RoBERTa-based GPT-2 classifiers\footnote{\url{https://github.com/openai/gpt-2-output-dataset/tree/master/detector}} (Base, Large) and HC3 ChatGPT detector\footnote{\url{https://huggingface.co/Hello-SimpleAI/chatgpt-detector-roberta}}. HC3 is the latest corpus targeted for detecting ChatGPT-generated texts \cite{guo2023close}.
We employ default parameters for each detection method.\footnote{To adopt DetectGPT, we change only $\mathsf{buffer\_size}$ from default to 2 in order to escape being stuck in the perturbation step.}

\paragraph{Attacking Methods}
Our OUTFOX attacker is based on ChatGPT (gpt-3.5-turbo-0301).
In generating attacks, we configure the $\mathsf{temperature}$ parameter to 1.3.
Our attacker takes retrieved in-context examples for a problem statement. The number of in-context examples is 10.
We compare our attacker to the paraphrasing attack by the DIPPER (as explained \S\ref{attacking}). The DIPPER paraphrases the non-attacked essays for attacking.
We configure both parameters for vocabulary $L$ and content re-ordering $O$ to 60, which are the parameters found to produce the strongest attack in \citet{krishna2023paraphrasing}. Other hyperparameters are set to the defaults.

\label{results}
\subsection{Results}
\paragraph{How Robust Is Our Detector, considering Attacks, against Attacked Texts?}
Table \ref{robustness_of_our_detector} presents the difference in detection performance of our OUTFOX detector with and without considering attacks on attacked essays. The attacking models include our OUTFOX attacker and the DIPPER.
Throughout all attackers, our detector improves the detection performance when considering attacks.
For instance, on our attacker-generated essays, our detector shows +41.3 points F1-score and +21.6 points AvgRec improvements when considering our attack. 
From this result, we empirically observe that our detector learns to detect essays from attackers via in-context examples. 
Notably, our detector, when considering our attack, shows the performance improvements on any attacked essay, while our detector considering the DIPPER improves only on the attacked essays by the DIPPER, but not our attacker.
This observation suggests that our attacker is not merely paraphrasing but may generate semantically diverse essays from given problem statements. Consequently, our detector is enabled to identify texts that employ a broader spectrum of attacks.

\paragraph{Does Our Detector, considering Attacks, Consistently Perform Well Even on Non-attacked Texts?}
Table \ref{side_effect} shows the difference in the detection performance of our OUTFOX detector on non-attacked essays, with and without considering attacks.
Our detector consistently performs well, even on non-attacked essays. The difference is minimal: an average decrease of only a -0.1 point F1-score and a -0.32 point AvgRec across all comparisons.
Furthermore, in non-attacked essays by FLAN-T5-XXL, our detector performs better than without considering attacks.
These results empirically show that considering attacks has little negative effects on the detection performance of our detector on the non-attacked texts.

\begin{table*}[t]
\centering
\small
\setlength{\tabcolsep}{6pt} 
\renewcommand{\arraystretch}{1.05} 
\begin{tabular}{ccccccc}\hline
\multirow{2}{*}{\textbf{Baseline type}}&\multirow{2}{*}{\textbf{Essay Generator}}& \multirow{2}{*}{\textbf{Detector}} & \multicolumn{4}{c}{\textbf{Metrics (\%) ↑}}\\\cline{4-7}
& & & \textbf{HumanRec} & \textbf{MachineRec} & \textbf{AvgRec} & \textbf{F1}\\\hline
\multirow{6}{*}{Statistical outlier methods}&\multirow{6}{*}{FLAN-T5-XXL}&\multicolumn{1}{l}{log $p(x)$} & 2.0 & 97.6 & 49.8 & 66.0\\
& &\multicolumn{1}{l}{Rank} & 28.8 & 86.2 & 57.5 & 67.0 \\
& &\multicolumn{1}{l}{LogRank} & 12.0 & 90.6 & 51.3 & 65.0\\
& &\multicolumn{1}{l}{Entropy} & 39.4 & 80.4 & 59.9 & 66.7\\
& &\multicolumn{1}{l}{DetectGPT} & 29.8 & 76.2 & 53.0 & 61.9\\
& &\multicolumn{1}{l}{OUTFOX} & 97.0 & 73.4 & \textbf{85.2} & \textbf{83.2}\\\hline
\multirow{8}{*}{Supervised classifiers}&\multirow{4}{*}{ChatGPT}&\multicolumn{1}{l}{RoBERTa-base} & 93.8 & 92.2 & 93.0 & 92.9\\
& &\multicolumn{1}{l}{RoBERTa-large} & 91.6 & 90.0 & 90.8 & 90.7 \\
& &\multicolumn{1}{l}{HC3 detector} & 79.2 & 70.6 & 74.9 & 73.8 \\
& &\multicolumn{1}{l}{OUTFOX} & 97.8 & 92.4 &\textbf{95.1} & \textbf{95.0}\\
\cline{2-7}
& \multirow{4}{*}{GPT-3.5}&\multicolumn{1}{l}{RoBERTa-base} & 93.8 & 92.0 & 92.9 & 92.8\\
& &\multicolumn{1}{l}{RoBERTa-large} & 92.6 & 92.0 & 92.3 & 92.3 \\
& &\multicolumn{1}{l}{HC3 detector} & 79.2 & 85.0 & 82.1 & 82.6 \\
& &\multicolumn{1}{l}{OUTFOX} & 97.6 & 96.2 & \textbf{96.9} & \textbf{96.9}\\
\hline
\end{tabular}
\caption{
Comparison of the detection performances of our OUTFOX detector and prior approaches on non-attacked essays. Prior approaches include statistical outlier detectors and supervised classifiers. We compare our detector with statistical outlier detectors on the essays by FLAN-T5-XXL and supervised classifiers on the essays by ChatGPT and GPT-3.5.
}
\label{comparison_baseline}
\end{table*}

\paragraph{Is Our Attacker Stronger than the Previous Paraphrasing Attack Approach?}
Table \ref{our_stronger_attacker} provides the detection performance on attacked essays by different attacking approaches: our attack and the DIPPER attack. Here, our detector does not consider any attacks.
Our attacker drastically degrades the detection performance of our detector and supervised classifiers the most by up to -57.0 points F1-score and -69.2 points MachineRec. 
In addition, our attacker has a more detrimental impact on the detection performance of all detectors than the DIPPER by up to -39.6 points F1-score.
We also find that the DIPPER doesn't degrade the detection performance much and conversely improves the detection performance, especially on RoBERTa-large. This is partially because the DIPPER attack is based on paraphrasing and not designed specifically to attack detectors resulting in the opposite effect.

\section{Comparison with Prior Work}
\label{comparison}
In this section, we compare the detection performance of our OUTFOX detector, when considering our attack, and prior detectors on non-attacked essays.
The prior detectors are divided into two groups: statistical outlier approaches and supervised classifiers.

\paragraph{Statistical Outlier Approaches}
Statistical outlier approaches need access to model logits for their detection.
Thus, we contrast our detector with these statistical outlier approaches in detecting essays by FLAN-T5-XXL.
As shown in Table \ref{comparison_baseline}, our detector has a far superior detection performance of 80.5 points F1-score and 83.5 points AvgRec to previous statistical approaches. 
We find that statistical outlier approaches tend to aggressively label Human-written essays as LLM-generated from the contrast between the low HumanRec and the high MachineRec.
While the HumanRec of our detector is high: 98.8 points, partially because our detector is based on ChatGPT, which is trained with human feedback, thus avoiding aggressive detection.

\paragraph{Supervised Classifiers}
We compare our detector and supervised classifiers in detecting essays generated by each ChatGPT and GPT-3.5.
Table \ref{comparison_baseline} shows that our detector has better detection performance on both essays by ChatGPT and GPT-3.5 than supervised classifiers.
We empirically find that ChatGPT has the few-shot ability to detect LLM-generated texts via labeled in-context examples without any parameter updates.

In summary, our OUTFOX detector shows the state-of-the-art detection performance on non-attacked essays.

\begin{figure}[t]
  \centering
  \includegraphics[width=\columnwidth]{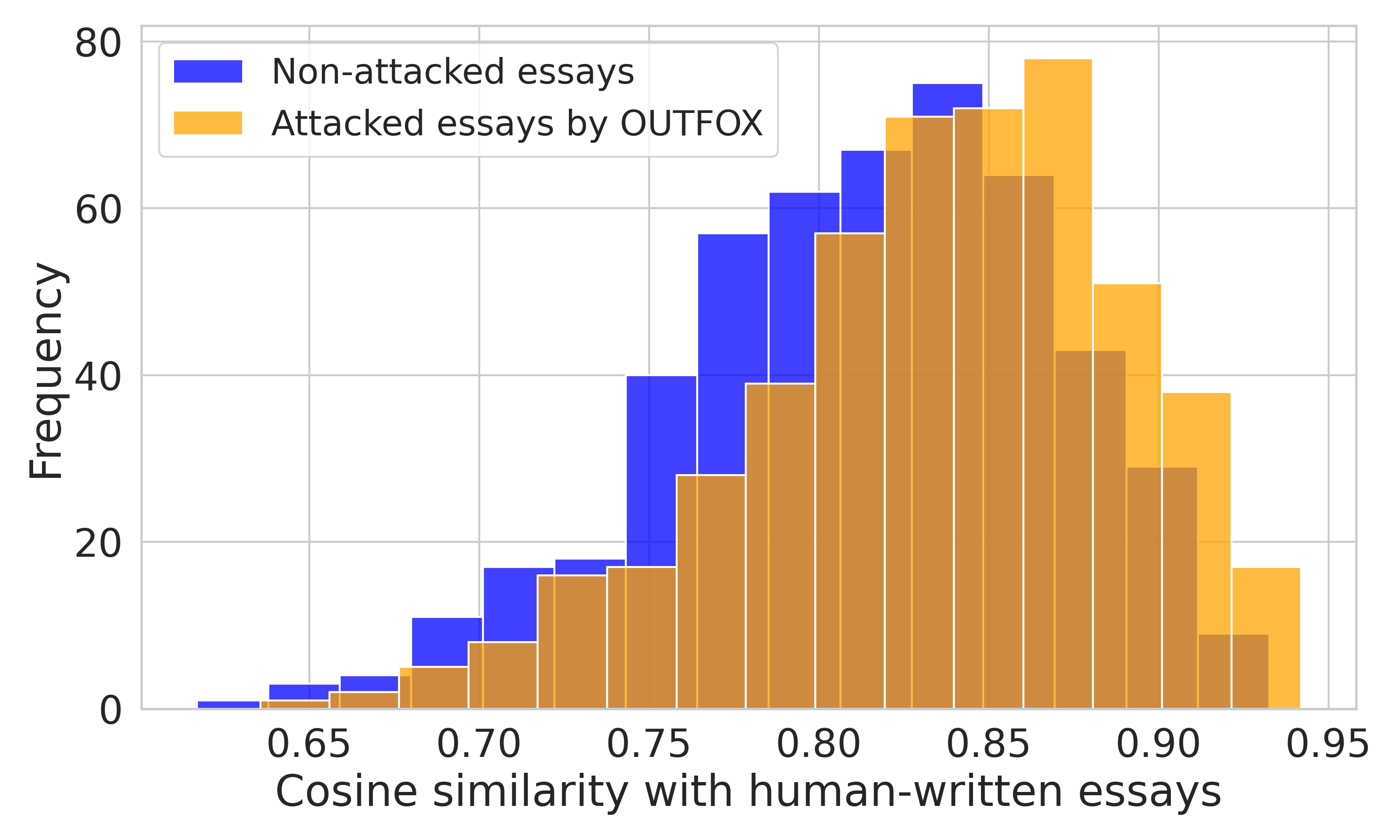}
  \caption{Cosine similarity distributions of non-attacked essays and our OUTFOX attacker-generated essays with human-written essays, respectively.}
  \label{fig:analysis}
\end{figure}

\section{The Semantic Similarity of Our OUTFOX Attacks with Human-Written Essays}
\label{analysis}
To explore why our attacker substantially degrades the detectors' performance, we investigate the difference between the semantic similarity of non-attacked essays and our attacker-generated essays with human-written essays.
We focus on our test set, consisting of 500 human-written, 500 ChatGPT-generated, and 500 attacked essays generated by our attacker.
For determining semantic similarity between two essays, we employ a pre-trained BERT model\footnote{\url{https://huggingface.co/bert-large-cased}} to each essay and compute a cosine similarity of the resulting embeddings.
Figure \ref{fig:analysis} presents cosine similarity distributions of non-attacked and our attacker-generated essays with human-written essays.
A rightward shift implies that our attacker-generated essays are more semantically similar to human-written essays than non-attacked essays.
From the shift, we empirically find that our OUTFOX attacker can generate essays that are more semantically similar to human-written essays than the non-attacking normal LLM, leading to difficulty in detecting attacker-generated essays.

\section{Conclusion}
We proposed OUTFOX, a framework that improves the robustness of the detector against attacks by allowing both the detector and the attacker to consider each other's outputs as examples for in-context learning.
The experiments in the domain of student essays demonstrate that 1) Our detector can learn to detect essays from attackers via in-context examples and 2) Notably, considering attacks of our detector has little negative effect on the detection of non-attacked texts. and 3) Our attacker, which is designed specifically to deceive the detector, can evade current LLM-generated text detectors more effectively than the previous paraphrasing attack. Furthermore, our analysis reveals that our attacker can generate an essay that is semantically closer to a human-written essay than a non-attacked essay, leading to success in effective attacking.
In future work, we will apply our framework to other domains, such as fake news generation and academic paper writing.



\section*{Acknowledgements}
These research results were obtained from the commissioned research (No.22501) by National Institute of Information and Communications Technology (NICT), Japan.

\bibliography{aaai24}
\clearpage

\end{document}